%% file: main.tex
\documentclass{article}
\usepackage{arxiv}
\usepackage[utf8]{inputenc}
\usepackage[T1]{fontenc}

\usepackage{microtype}
\usepackage{amsmath,amssymb,booktabs,graphicx,array}
\usepackage{float,placeins}
\usepackage[numbers,sort&compress]{natbib}
\usepackage{xurl}
\usepackage{xcolor}
\usepackage[colorlinks=true,linkcolor=blue!45!black,citecolor=blue!45!black,urlcolor=blue!45!black]{hyperref}
\usepackage[font=small,labelfont=bf]{caption}
\hypersetup{pdftitle={Can Jev Judge Radiology Reports? Evaluating a System One Model for Clinical Factuality},pdfsubject={Radiology report evaluation with Jev},pdfkeywords={Jev, System One, radiology, factuality, evaluation}}
\hypersetup{pdfauthor={Jiaju Huang, Hao Yang, Xinyu Ma, Xinglong Liang, Kunyan Cai, Junqiang Ma, Shaobin Chen, Yue Sun, Tao Tan}}
\renewcommand{\shorttitle}{Can Jev Judge Radiology Reports?}
\title{Can Jev Judge Radiology Reports?\\
Evaluating a System One Model for Clinical Factuality}
\author{\parbox{0.96\textwidth}{\centering\normalfont
Jiaju Huang\textsuperscript{1}, Hao Yang\textsuperscript{1},
Xinyu Ma\textsuperscript{1}, Xinglong Liang\textsuperscript{2,3},\\
Kunyan Cai\textsuperscript{1}, Junqiang Ma\textsuperscript{1}, Shaobin Chen\textsuperscript{1},
Yue Sun\textsuperscript{1}, Tao Tan\textsuperscript{1,*}\\[6pt]
\small
\textsuperscript{1}Intelligent Medical Computing Laboratory, Faculty of Applied Sciences,\\
Macao Polytechnic University\\[3pt]
\textsuperscript{2}Department of Radiology, Netherlands Cancer Institute,\\
Amsterdam, the Netherlands\\[3pt]
\textsuperscript{3}Department of Radiology and Nuclear Medicine,\\
Radboud University Medical Center, Nijmegen, the Netherlands\\[6pt]
\textsuperscript{*}Corresponding author: Tao Tan,
\href{mailto:taotanjs@gmail.com}{\texttt{taotanjs@gmail.com}}
}}
\date{September 23, 2026}
\begin{document}
\maketitle
\input{sections/00_abstract}
\input{sections/01_intro}
\input{sections/02_related}
\input{sections/03_method}
\input{sections/04_setup}
\input{sections/05_results}
\input{sections/06_discussion}
\input{sections/07_conclusion}
\input{sections/10_availability}
\FloatBarrier
\begingroup
\small
\setlength{\bibsep}{0pt}
\bibliographystyle{plainnat}
\bibliography{refs}
\endgroup
\clearpage
\appendix
\small
\setcounter{table}{0}
\setcounter{figure}{0}
\renewcommand{\thetable}{A\arabic{table}}
\renewcommand{\thefigure}{A\arabic{figure}}
\input{sections/08_appendix}
\end{document}

%% file: sections/00_abstract.tex
\begin{abstract}
An AI-generated radiology report can resemble a physician's report while
omitting an abnormality, adding an unsupported finding, or reversing its
presence. Measuring these factual differences is essential for evaluating
report generators. We study Jev, a System One decision model, as a simple,
low-cost judge of agreement with physician-written reference reports.
Our evaluator checks whether each statement is supported by the other report
and combines these judgments in both directions to capture unsupported claims
and omissions. A single-question configuration reaches Kendall correlations
of 0.573 on RadEvalX and 0.398 on RadEvalExpert with expert error counts,
outperforming an open natural language inference judge under matched
decomposition and aggregation. One support question per statement retains
similar expert agreement to seven while using 43--45\% fewer judgment input
tokens. Judgment API calls cost less than US\$0.03 per 100 report pairs, excluding
local decomposition.
In a separate controlled-error test, Jev detects false negation with an AUROC
of 0.977. Local RadMatch achieves stronger agreement on clinically significant
errors in both expert datasets and on total errors in the shared RadEvalExpert
subset. Finding-count and error-scope analyses show that benchmark agreement
reflects report size and error definitions as well as medical error detection.
These results support Jev as a practical judgment component for measuring
factual differences in generated radiology reports and identify where more
elaborate evaluation remains valuable.
\end{abstract}

%% file: sections/01_intro.tex
\section{Introduction}
A generated radiology report can read naturally and still change the medical
story. It may add a finding that the physician's report never mentions,
leave out an abnormality, or describe it on the wrong side. Changing
``a pleural effusion'' to ``no pleural effusion'' preserves most words while
reversing the finding. For researchers developing report generators, these
are the differences an evaluator needs to capture: which assertions disagree
with the reference, and which findings have been left out?

This task calls for many small judgments. For each statement, does the
reference support it, contradict it, or leave it unaddressed? Checking in the
reverse direction asks whether the generated report covers the reference's
findings. Jev offers a direct interface for making these decisions: its
System One model returns typed answers with probabilities that software can
combine into a report score~\citep{jev}. We ask whether Jev can measure factual
differences in generated radiology reports with just one support question
per statement.

We study Jev as a reference-based report evaluator. We extract statements
from both reports, judge each against the other report, and aggregate their
discrepancies (Figure~\ref{fig:method}). This design builds on report
decomposition and bidirectional entailment, particularly RadFact~\citep{radfact}.
We examine what Jev contributes within this framework and how much complexity
is needed to use it effectively. Expert-annotated reports test whether the
scores reflect overall error burden; controlled errors reveal which medical
changes the judge detects. Component comparisons, token costs, and repeated
calls test whether the evaluator is economical and stable enough for repeated
report-generator comparisons.

The results support a simple configuration. One support question per statement
produces expert agreement close to that of the seven-question design while
reducing judgment API input cost by 43--45\%. The resulting judgments cost
under three cents per hundred report pairs, excluding local decomposition.
Under matched decomposition and aggregation, this configuration outperforms
an open natural language inference (NLI) judge. In a separate sentence-level
test, Jev detects false negation---turning a present finding into an absent
one---with an AUROC of 0.977. Its report rankings also repeat closely across
five runs. Local RadMatch achieves higher agreement on clinically significant
errors in both expert datasets and on total errors in the shared RadEvalExpert
subset, showing where its more elaborate procedure remains useful.

A report that describes more findings offers more opportunities for errors.
An evaluator can therefore correlate with expert error counts by capturing
report size, even without identifying each error correctly. Jev
exceeds a finding-count baseline in both expert datasets. We complement this
comparison with candidate ordering within the same examination and with a
separation of factual errors from spelling and repetition perturbations.
The latter changes the relative ranking of Jev and NLI.

Our contributions are threefold. First, we evaluate Jev's ability to measure
factual discrepancies between generated and reference radiology reports,
using expert error counts and controlled medical changes. Second, we identify
a single-question configuration that combines competitive expert agreement
with low judgment cost. Third, we characterize what the evaluator
captures and how it behaves through analyses of report size, error scope,
calibration, and repeatability.

%% file: sections/02_related.tex
\section{Related Work}
\paragraph{Radiology report metrics.}
Report evaluation spans lexical overlap, contextual similarity, extracted
clinical structure, and learned error scores. BERTScore compares contextual
token representations~\citep{bertscore}; RadGraph measures agreement in
clinical entities and relations~\citep{radgraph}; RaTEScore uses radiology
entities and their attributes~\citep{ratescore}. RadCliQ combines metric
signals to predict expert error counts~\citep{radcliq}. The RadEval framework
provides common implementations and an expert-annotated evaluation
dataset~\citep{radeval}. These methods differ in their targets: a similarity
ratio and a predicted number of errors need not rank reports identically.

\paragraph{Language models as report judges.}
GREEN identifies and describes clinically meaningful discrepancies between
reports~\citep{green}. RadFact decomposes reports into single-finding phrases
and verifies entailment in both directions, yielding logical precision and
recall~\citep{radfact}. Its evidence extraction also supports grounded
evaluation. RadMatch uses finding-level matching and attribute analysis to
produce auditable discrepancy measures~\citep{radmatch}. Our work adopts the
established decomposition-and-entailment structure and examines Jev as a
judgment engine. CRIMSON adds context-sensitive clinical significance
and radiologist preference benchmarks~\citep{crimson}. AtomiMed separates
disease and attribute facts and verifies them through bidirectional
questions across imaging modalities~\citep{atomimed}.

Recent work also examines how to configure medical judges.
VERT compares judge models, prompts, reasoning settings, ensembling, and
fine-tuning on RadEval and RaTE-Eval~\citep{vert}. Its analysis also discusses
how finding counts affect normalized scores. RadSEM combines atomic findings,
contradiction-constrained matching, and deterministic weighted scoring~\citep{radsem}.
It evaluates sensitivity to graded semantic corruption and synonym--antonym
rewrites. We focus on Jev's typed probability outputs, its medical error
detection, and the cost of a minimal question set. Our count baseline and
error-scope comparisons examine what expert-agreement scores reward.

\paragraph{Evaluating the evaluators.}
General LLM-judge studies document position and verbosity biases alongside
agreement with human preferences~\citep{mtbench}. Length-controlled AlpacaEval
uses regression adjustments to reduce preference for longer responses~\citep{lengthcontrol}.
In summarization, SummEval includes output length among its simple comparison
features~\citep{summeval}. Radiology benchmarks add a related measurement issue:
reports with more findings offer more opportunities for counted errors.
RaTE-Eval already normalizes sentence errors by the number of potential
errors~\citep{ratescore}. We measure the strength of a finding-count baseline
on expert report benchmarks and contrast pooled error burden with candidate
ordering within the same examination.

\paragraph{Probability quality and benchmark interpretation.}
Probabilistic outputs allow soft scores and task-specific calibration, but
probability values alone do not establish calibration~\citep{calibration}.
We compare Jev's raw probabilities with those of a public DeBERTa NLI
checkpoint~\citep{nli}, and evaluate post-hoc calibration on held-out studies.
Expert datasets such as ReXVal~\citep{rexval}, RadEvalX~\citep{radevalx_iu},
and RadEvalExpert~\citep{radeval}
measure report-level agreement, whereas ReXErr~\citep{rexerr} provides
controlled local errors. We use both forms of evidence and test how simple
count signals and error definitions affect their conclusions.

%% file: sections/03_method.tex
\section{Jev as a Radiology Report Judge}
\label{sec:method}
\subsection{Task and statement units}
Given a generated report $C$ and a physician-written reference report $R$,
we assess which findings disagree and which are left unaddressed.
Let $c_1,\ldots,c_n$ and $r_1,\ldots,r_m$ denote the
statements extracted from the two reports.

We implement two granularities. L0 applies sentence segmentation and removes
pure section headings. L1 uses a local Qwen3.5-9B model to split each sentence
into atomic findings while preserving negation, uncertainty, laterality,
severity, measurements, and comparison phrases. Invalid JSON falls back to
the original sentence. The reported OneQ configuration uses L1. L0 is a
separate ablation.

\begin{figure}[t]
\centering\includegraphics[width=\linewidth]{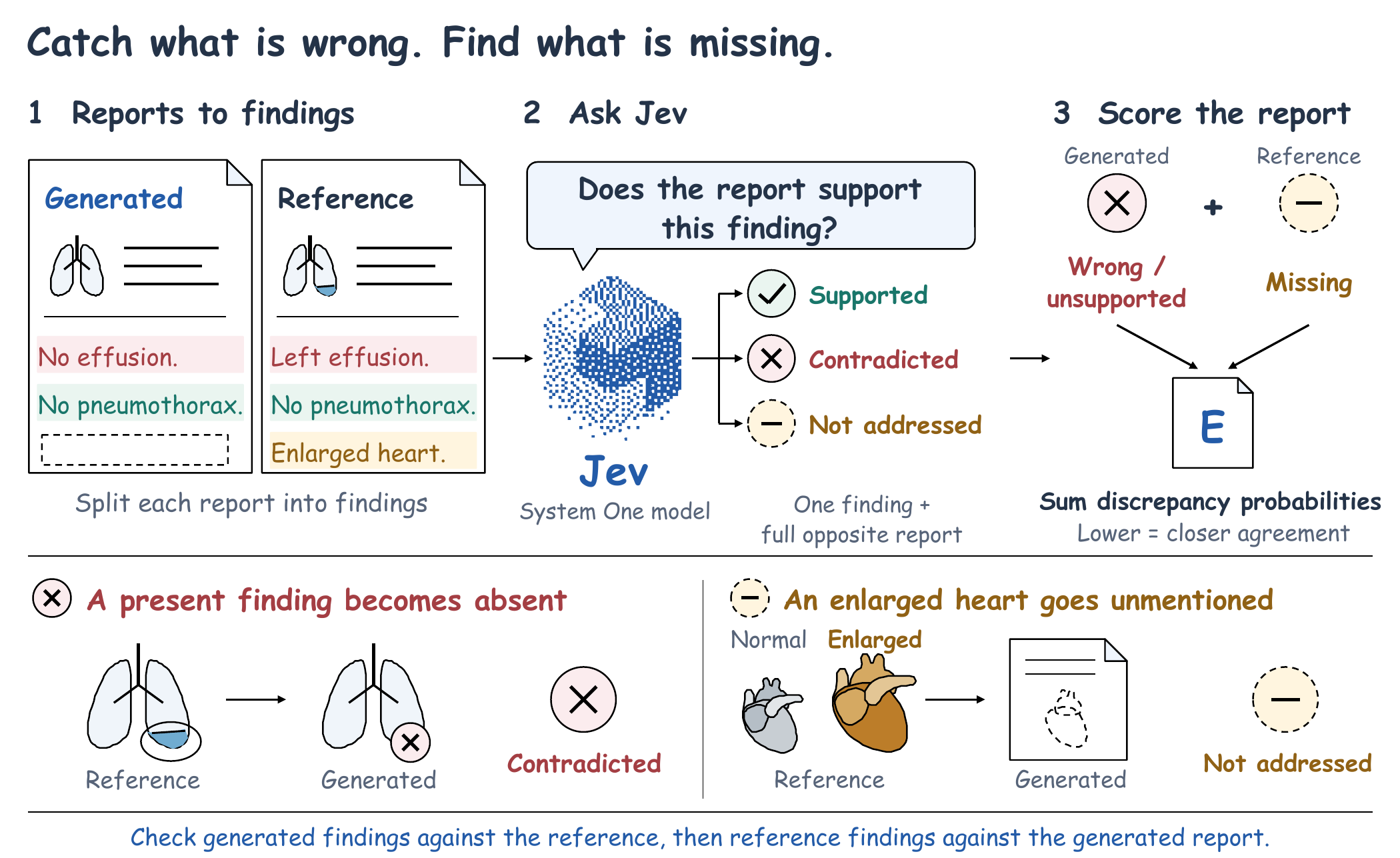}
\caption{\textbf{Jev checks what a report gets wrong and what it leaves out.}
OneQ compares each finding with the complete opposite report in both directions.
The constructed example visualizes a denied effusion and an omitted enlarged
heart; a normal-sized heart is shown for visual comparison. Jev returns support,
contradiction, and noncoverage probabilities. Scores sum candidate-side
contradiction/noncoverage with reference-side noncoverage
(Eq.~\ref{eq:error}); lower is better.
Model illustration adapted from \href{https://typesafe.ai}{TypeSafe AI}.}
\label{fig:method}
\end{figure}

\subsection{One question, three outcomes}
For a statement $x$ and an opposing report $Y$, Jev returns
\begin{equation}
\bigl(p_s(x,Y),p_c(x,Y),p_u(x,Y)\bigr),
\qquad p_s+p_c+p_u\approx 1,
\end{equation}
for \emph{supported}, \emph{contradicted}, and \emph{not addressed}.
We use the returned probabilities at API precision without renormalization.
The prompt asks whether the report supports the claim about an imaging
finding. Its criteria distinguish compatible descriptions, incompatible
presence or attributes, and findings absent from the report text.
For example, ``left pleural effusion'' is contradicted by an explicit denial
of left pleural effusion; a report that never addresses effusion supplies no
support for the claim. Here, \emph{not addressed} denotes missing textual evidence.

The same question is applied to every candidate statement against $R$ and
every reference statement against $C$. OneQ therefore means one question per
statement--report input; a report pair requires multiple such judgments.

\subsection{Bidirectional aggregation}
Our primary score sums candidate-side discrepancies and reference-side
noncoverage:
\begin{equation}
 E(C,R)=\sum_{j=1}^{n}\left[p_c(c_j,R)+p_u(c_j,R)\right]
       +\sum_{i=1}^{m}p_u(r_i,C).
\label{eq:error}
\end{equation}
Lower values indicate greater agreement. The asymmetry avoids adding a
second contradiction term from the reference direction. This is a soft
discrepancy count over extracted statements; its numerical agreement with
expert error counts is evaluated separately from its ranking ability.

We also report soft precision, recall, and F1:
\begin{equation}
 P=\frac{1}{n}\sum_j p_s(c_j,R),\qquad
 R_s=\frac{1}{m}\sum_i p_s(r_i,C),\qquad
 F_s=\frac{2PR_s}{P+R_s}.
\label{eq:f1}
\end{equation}
The normalized count is $E/(n+m)$. Empty-side means and undefined harmonic
means are set to zero in the implementation. We report counts against count
labels and use the normalized and F1 variants to examine agreement as a
proportion of report content.

\subsection{Configurations and open judge}
\textbf{OneQ} uses only the support question. \textbf{Full} adds four
negation/uncertainty questions, a discrepancy-type question, and a clinical
significance question. Full also applies a negation-consistency correction:
if the claim and report negation probabilities differ by more than 0.5 while
$p_s>0.5$, half of $p_s$ is moved to $p_c$. OneQ has no negation outputs, so
this correction is inactive. Significance and error-type outputs enter their
own variants and do not otherwise alter Eq.~\ref{eq:error}.

We test sentence-only splitting, shorter criteria, relevance filtering,
alternative discrepancy weights, and significance weighting. Relevance
filtering asks whether each report sentence concerns the claim's anatomical
entity and retains probabilities above 0.3. An empty selection yields an
empty opposing report. We evaluate each component separately against Full.

The open judge uses
\texttt{DeBERTa-v3-large-mnli-fever-anli-ling-wanli}~\citep{nli}.
The opposing report is the premise and the statement is the hypothesis.
Entailment, contradiction, and neutral probabilities map to $p_s,p_c,p_u$.
It shares L1 statements and Eq.~\ref{eq:error} with OneQ; its tokenizer uses
a maximum sequence length of 512 tokens. This comparison isolates the judge
within the evaluated configuration, including its input handling.

%% file: sections/04_setup.tex
\section{Experimental Design}
\subsection{Datasets and targets}
\textbf{Expert agreement.} RadEvalX contains 100 IU-Xray reference reports,
each paired with one M2Tr-generated report~\citep{radevalx_iu,iuxray}.
Two radiologists reached consensus on significant and insignificant error
counts across eight categories. The sample contains 80 abnormal and 20 normal
reports, selected after stratification by RadCliQ scores.
We evaluate total and clinically significant error counts over all eight
categories; a six-category sensitivity analysis excludes the two uncertainty
categories (Appendix~\ref{app:data}). RadEvalExpert contains 624 pairs from 208 studies, with three
model-generated candidates per reference~\citep{radeval}. It covers findings
and impression sections and supplies total and significant error labels.
We keep each study together during uncertainty estimation.

\textbf{Controlled errors.} The ReXErr sentence test set contains 19,859
examples~\citep{rexerr}. Excluding 346 neutral rewrites leaves 19,513 examples:
8,723 injected-error sentences and 10,790 unchanged sentences. Each candidate
sentence is judged against its complete original report. This experiment uses
the raw support probabilities from the Full question set, without atomic
decomposition or negation correction. It tests sentence-level detection and
calibration separately from the report-level OneQ pipeline. We retain all
11 positive error types and also analyze eight factual-error types and three
language-perturbation types (typos, homophones, and repetition), grouped by
the dataset's intervention labels.

We use 2,000 sampled ReXErr report pairs for cost and report-level diagnostics.
RaTE-Eval supplies supplementary sentence, paragraph, and synthetic comparison
tests~\citep{ratescore}. Their targets and processing are described in
Appendix~\ref{app:data}.

\subsection{Comparators and implementation}
\label{sec:implementation}
The main comparators include GREEN, RadCliQ, RadGraph F1, BERTScore, Temporal
F1, open NLI, and the finding-count baseline $n+m$. RadEval supplies the
standard metric implementations~\citep{radeval}. We also run RadFact and
RadMatch on all 100 RadEvalX pairs and a fixed random subset of 200 RadEvalExpert
pairs. All subset comparisons use exactly the same valid pairs.

Both generative baselines use a local Qwen3.8-27B-FP8 backend
on two 32\,GB RTX 5090 D GPUs. This replaces the original
papers' judge choices. On the RadEvalExpert subset, RadMatch retried 52 pairs
and used validation fallback on 11; its default backend configuration sampled
at temperature 1.0.
RadFact used temperature zero.
On RadEvalX, RadMatch used validation fallback on 8 of 100 pairs; all pairs
received a final score. Both baselines disabled thinking. RadFact used an
8,192-token context, and RadMatch used 16,384 tokens to accommodate the
complete few-shot prompt, with a 4,096-token output cap.
GREEN's default decoding is greedy.

Our implementation, S1Fact, uses Jev version
\texttt{jev-1.13.0}. L1 uses
temperature-zero local inference. Repeated-call experiments use fresh
judgments with decomposition held fixed. Additional prompt and implementation details appear in
Appendix~\ref{app:implementation}.

\textbf{Repeatability.} On a fixed random subset of 50 RadEvalX pairs,
we repeat both Full and OneQ five times, holding decomposition fixed
and bypassing the judgment cache.
A matched experiment uses a fixed selection of 50 pairs from the
200-pair RadEvalExpert subset, covering 48 studies. We run Jev--Full and
Jev--OneQ five times each on the same 616 extracted findings, bypassing the
judgment cache. Five runs per baseline setting evaluate GREEN with greedy
decoding, RadFact at temperature zero, and RadMatch at temperatures zero and
one. The latter is the local backend's default. Jev holds L1 decomposition
fixed; the baselines repeat their local evaluation procedures.
We measure exact score agreement, between-run ranking consistency, and
variation in correlation with expert total-error counts.

\subsection{Statistics and cost}
Primary agreement is Kendall's $\tau_b$, oriented in advance so higher is
better: error scores retain their sign against error labels, and similarity
scores have their sign reversed. We do not take absolute correlations.
Confidence intervals use percentile bootstrap over studies: 2,000 resamples
for RadEvalX and 1,000 resamples for RadEvalExpert. Pairwise
comparisons first select rows where both methods and the target are finite,
then resample the same studies for both methods. Intervals describe individual
comparisons and are not adjusted for multiple testing.

On RadEvalExpert, which supplies multiple candidates per study, we separately
compute within-study ordering,
\begin{equation}
 W=\frac{N_{\mathrm{concordant}}-N_{\mathrm{discordant}}}
 {N_{\mathrm{all\ within\mbox{-}study\ candidate\ pairs}}}.
\label{eq:within}
\end{equation}
Ties contribute zero to the numerator and remain in the denominator. This
statistic is not tie-corrected $\tau_b$. It holds the reference report fixed
within each comparison, while candidate lengths may still differ. Bootstrap
copies of a study contribute independent copies of its pair counts.

Controlled-error analyses report AUROC, recall at 0.5, ten-bin expected
calibration error (ECE), and Brier score. Error-scope intervals use
400 study resamples. Post-hoc logistic and isotonic calibration fit on one half
of the studies and are evaluated on the other half. Raw and calibrated
probabilities are compared on the same holdout.

We calculate API cost from input-token usage at \$0.042 per million tokens;
output tokens are free~\citep{jev}.
Configuration comparisons count every judgment at that price, including
judgments served from the local cache. This measures the cost of the complete
judgment workload at a common price. Local decomposition
cost is separate. We report request latency, per-report judgment time,
and batch completion time separately.

%% file: sections/05_results.tex
\section{Results}
We first test whether report scores reflect expert-assessed factual errors
and which medical changes Jev detects. We then examine the complexity and
cost needed for these judgments, followed by analyses of score interpretation,
calibration, and repeatability.

\subsection{Report scores track expert-assessed factual differences}
\label{sec:agreement}
The report-level test asks whether greater disagreement with the reference
corresponds to more errors identified by radiologists.
Jev--OneQ reaches $\tau_b=0.573$ on RadEvalX and $0.398$ on RadEvalExpert
(Table~\ref{tab:main}). Full reaches 0.582 and 0.397, respectively.
With the same statements and count aggregation, OneQ improves over open NLI
by $0.236$ [0.131, 0.345] and $0.075$ [0.043, 0.106] on total errors.
Its advantage over NLI also holds for significant-error labels.
OneQ exceeds RadCliQ on total errors in both datasets. On RadEvalX,
its difference from GREEN is $0.125$ [$-0.006$, 0.256], while
significant-error correlation is nearly equal (0.346 versus 0.347).
On RadEvalExpert, OneQ exceeds both GREEN and RadCliQ on both targets.
Appendix~\ref{app:metrics} reports the complete standard metric suite.

\input{tables/table1}

RadMatch's advantage depends on the error target. On RadEvalX, its
total-error correlation is
0.580, close to OneQ's 0.573 and Full's 0.582. The paired OneQ--RadMatch
difference is $-0.007$ [$-0.137$, 0.114]. For significant errors, RadMatch
reaches 0.491 versus OneQ's 0.346; the difference is
$-0.145$ [$-0.294$, $-0.004$]. Full also trails RadMatch on this target.
Local RadFact reaches 0.477 on total errors and 0.398 on significant errors;
neither paired difference from OneQ excludes zero.

RadMatch leads on both error targets in the shared RadEvalExpert subset.
Across these 200 pairs, its total-error correlation is 0.453, compared with
0.322 for OneQ and 0.320 for Full. OneQ minus RadMatch is
$-0.131$ [$-0.226$, $-0.044$] for total errors and
$-0.226$ [$-0.320$, $-0.137$] for significant errors.
Both intervals exclude zero. On the same subset, OneQ exceeds local RadFact
for total errors by $0.171$ [0.052, 0.294]. Its significant-error difference
is $0.074$ [$-0.048$, 0.195].

\subsection{Detecting changes in medical assertions}
\label{sec:errors}
The sentence-level test asks whether Jev recognizes a specific change in
medical meaning. Using Full's raw support outputs in the ReXErr sentence
experiment, Jev detects false negation with an AUROC
of 0.977 [0.972, 0.982] and recall of 91.9\% at threshold 0.5.
It also detects changed locations (0.984), changed severity (0.975), and
added medical devices (0.989). These are concrete factual differences:
turning ``a left pleural effusion'' into ``no left pleural effusion'' reverses
presence, while changing ``left'' to ``right'' changes location.
Appendix~\ref{app:errors} reports all error types and a separate language
stress test.

\input{tables/table2}

The ranking changes when the test includes spelling and repetition
(Table~\ref{tab:scope}). Over all injected errors, NLI has higher AUROC
than Jev, 0.8969 versus 0.8896. Restricting positives to factual-error types
reverses that ordering: Jev reaches 0.9680 and NLI reaches 0.9614,
with a paired difference of $0.0066$ [0.004, 0.009].
For language perturbations, both scores fall, and NLI leads by 0.0674.
These 1,638 language perturbations constitute 18.8\% of all positive examples.

Jev's AUROC for added typos is 0.562, and its recall for repetition is 0.003.
Low sensitivity to meaning-preserving edits suits factuality evaluation.
Some language perturbations can also change medical meaning, so we interpret
this grouping by error type. Assessing overall writing quality requires
broader error coverage.

\subsection{One question retains agreement at lower cost}
\label{sec:ablation}
\begin{figure}[t]
\centering\includegraphics[width=\linewidth]{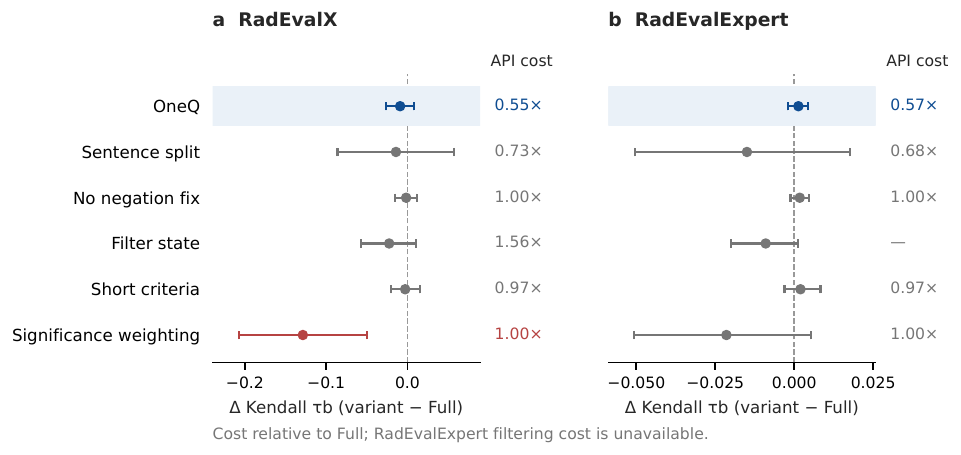}
\caption{Component changes relative to Full, evaluated independently.
Points show changes in total-error $\tau_b$; bars are paired 95\% study-bootstrap
intervals. API input cost uses the same price for all judgments, including
cached ones. RadEvalX filtering cost includes sentence-selection and
judgment calls; RadEvalExpert filtering cost is incomplete and omitted. Removing negation correction or changing aggregation reuses
the Full judgments and leaves their API cost unchanged.}
\label{fig:ablation}
\end{figure}

OneQ retains similar expert agreement with fewer input tokens
(Figure~\ref{fig:ablation}). Relative to Full, its total-error correlation
changes by $-0.009$ [$-0.026$, 0.008] on RadEvalX and
$+0.001$ [$-0.002$, 0.004] on RadEvalExpert.

The token savings are consistent across datasets. On RadEvalX, Full uses
9,695.17 input tokens per pair and OneQ uses 5,374.55, a 44.56\% reduction.
On RadEvalExpert, the corresponding means are 10,797.20 and 6,151.13,
a 43.03\% reduction. OneQ's judgment API cost is approximately
\$0.000226 and \$0.000258 per pair, excluding local decomposition.
In practical terms, the judgment API costs about 2.3--2.6 cents per hundred
report pairs.
Appendix~\ref{app:cost} gives the calculation and latency measurements.

Other additions bring no consistent improvement. Replacing atomic findings
with sentences changes correlation by $-0.014$ on RadEvalX and $-0.015$ on
RadEvalExpert; both intervals include zero. Removing negation correction or
shortening the support criteria also yields small changes.
On RadEvalX, filtering retains 32.2\% of report sentences and leaves 11.2\%
of statement inputs with an empty report. Correlation changes by $-0.023$
[$-0.057$, 0.011], while total judgment-and-filtering input cost rises by
56.1\%. On RadEvalExpert, the correlation change is about $-0.009$.
This filtering rule offers no observed scoring advantage.

Significance weighting reduces total-error correlation on RadEvalX by
$0.129$ [0.050, 0.208]. Its significant-error correlation rises from 0.343
to 0.396, but the paired interval for that gain includes zero.
On RadEvalX, jointly halving candidate and reference noncoverage weights
raises correlation by $0.053$ [0.005, 0.104]. This is an exploratory choice
from the nine-setting grid evaluated on the same labels; the main comparisons
retain the default weights.
On RadEvalExpert, halving candidate noncoverage weight raises correlation
by $0.013$ [$-0.006$, 0.035].
Appendix~\ref{app:ablation} reports the complete grid.
The practical outcome is a simple configuration: L1 statements, one support
question, and default count aggregation. The sentence-only and OneQ variants
were tested separately; their combined effect remains an open comparison.

\subsection{Judgment cost in a concrete example}
\label{sec:api-demo}
\begin{figure}[htbp]
\centering\includegraphics[width=\linewidth]{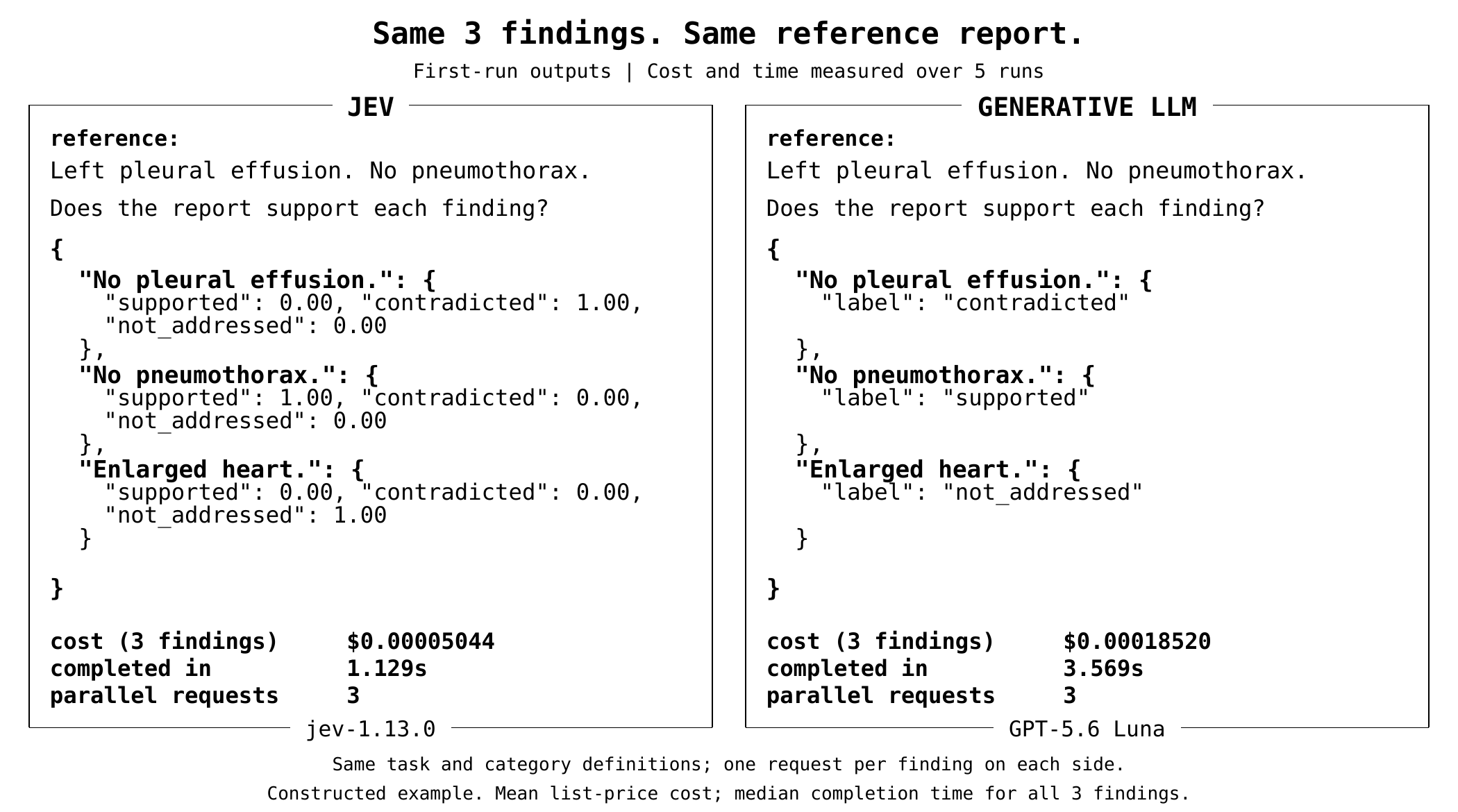}
\caption{Same findings, two ways to judge. Both methods receive the same
reference report, findings, and three-way support definitions. Outputs are
from the first run on this constructed example. Cost is the five-run mean
of token usage priced at official standard rates; time is the median wall
time to finish all three requests.}
\label{fig:api-demo}
\end{figure}

Figure~\ref{fig:api-demo} shows what a judgment call buys: a decision about
whether a report supports a finding. On three constructed examples covering
contradiction, support, and an unmentioned finding, Jev and GPT-5.6 Luna
return the same classifications in all five runs. Jev supplies the three
class probabilities directly; Luna generates a JSON label with reasoning disabled.
For the three findings together, mean API cost is
\$0.00005044 for Jev and \$0.00018520 for Luna. Median completion time is
1.129\,s and 3.569\,s, respectively, using three concurrent requests per method.
Appendix~\ref{app:api-demo} gives the protocol and pricing calculation.

\subsection{Interpreting agreement beyond finding counts}
\label{sec:length}
We compare Jev with a finding-count baseline to assess the contribution of
its support judgments beyond report size.
Counting extracted findings alone reaches $\tau_b=0.288$ on RadEvalX and
0.261 on RadEvalExpert. This baseline never checks whether a statement is
correct; it counts the opportunities for disagreement.
OneQ improves over it by $0.285$ [0.157, 0.416] and
$0.137$ [0.080, 0.197], respectively. Jev's judgments improve agreement
beyond the finding-count baseline.

\begin{figure}[t]
\centering\includegraphics[width=\linewidth]{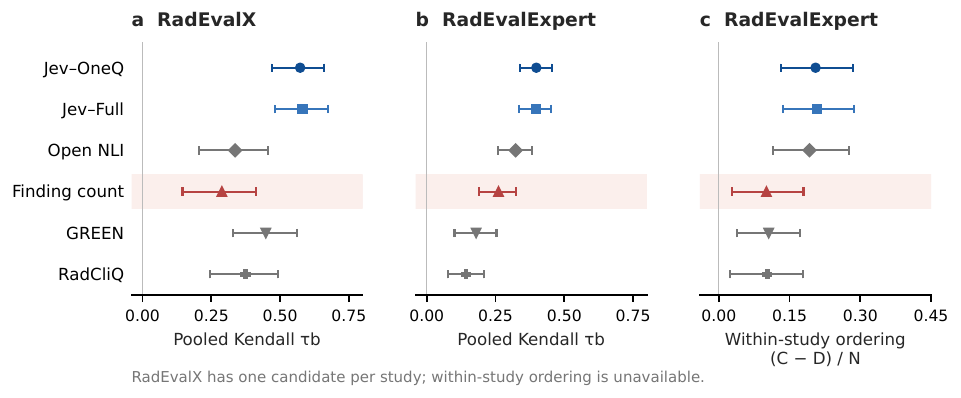}
\caption{Finding counts and the scope of comparison. Left and center:
pooled correlation with expert total-error counts on the two datasets.
Right: ordering of RadEvalExpert candidates for the same study, using
Eq.~\ref{eq:within}. All methods use the complete respective dataset.
Intervals resample studies. RadEvalX has only one candidate per study and
cannot support within-study ranking. The two statistics answer different
questions; within-study comparisons fix the reference report while candidate
lengths can still vary.}
\label{fig:length}
\end{figure}

On RadEvalExpert, OneQ also orders candidates describing the same examination
better than finding count in point estimate: 0.205 versus 0.101
(Figure~\ref{fig:length}). This view asks which candidate is better for a
particular examination, complementing the pooled ranking of error burden
across different examinations.

The common RadEvalExpert subset contains 69 within-study candidate
comparisons. Here, both Jev configurations score 0.246 for total errors,
compared with 0.043 for RadMatch. The paired difference is
$0.203$ [$-0.011$, 0.406]. For significant errors, the difference is
$0.043$ [$-0.169$, 0.250]. Both intervals include zero, leaving the
within-study difference between Jev and RadMatch uncertain.

Aggregation is another part of the explanation. On this same subset,
Full's count score reaches 0.320, whereas its soft F1 reaches 0.111.
The open NLI judge similarly changes from 0.294 with counts to 0.082 with F1.
Summation retains report-size information; F1 normalizes support within each
report. Both aggregation and judge choice influence agreement with error-count
labels. The size of the aggregation effect
also depends on the dataset: on RadEvalX, Full reaches 0.582 with counts and
0.561 with soft F1, a much smaller difference whose paired interval includes zero.

\subsection{Calibration makes probabilities useful for thresholding}
Jev's raw sentence-level probabilities have ECE 0.1192 on the full test
set, compared with 0.1045 for NLI. On the separate calibration holdout,
Jev's ECE falls from 0.1235 to 0.0077 after isotonic calibration.
NLI improves from 0.1072 to 0.0063 under the same procedure.
These results support task-specific calibration before using probability
thresholds. On RadEvalX, Full's highest score bin averages 8.53,
while the corresponding expert count averages 5.30.
Full tracks error burden, but its raw score overestimates expert error counts
at the upper end (Appendix~\ref{app:calibration}).

\subsection{Rankings remain stable across repeated judgments}
\label{sec:repeatability}
Jev's report rankings repeat closely in both configurations.
On the same 50 RadEvalExpert pairs used for the baselines, five uncached runs
give median between-run rank agreement of 0.990 for OneQ and 0.986 for Full
(Table~\ref{tab:repeatability}). The corresponding standard deviations in
expert correlation are 0.00381 and 0.00126.
Every report pair has some score variation across the five runs; the resulting
rankings remain close. Median score SD is 0.042 for OneQ and 0.047 for Full.
On 50 independently selected RadEvalX pairs, five runs each give correlation
SDs of 0.00745 for Full and 0.00737 for OneQ.
Appendix~\ref{app:repeatability} details the protocols and score variation.

The baseline runs show how much inference settings can change this
behavior (Table~\ref{tab:repeatability}). With the local backend's default
sampling, RadMatch gives identical scores across five runs for only 10\%
of report pairs. Setting temperature to zero raises that proportion to 82\%
and between-run rank consistency from 0.662 to 0.965.
The standard deviation of expert correlation falls from 0.03582 to 0.00975.
On this subset, mean expert correlation also rises from 0.400 to 0.468.

\input{tables/repeatability}

GREEN gives exactly the same scores in all five runs. RadFact's default
temperature-zero setup gives identical F1 scores for 88\% of pairs.
Repeatability varies across these pipelines and their decoding settings.

%% file: tables/table1.tex
\begin{table}[t]
\centering\small
\caption{Agreement with expert error counts. Signs are oriented so higher is better; brackets are 95\% study-bootstrap intervals. Panel A uses all pairs in each dataset for the first nine rows; local RadFact and RadMatch also cover all 100 RadEvalX pairs. Panel B restricts every method to the same 200 RadEvalExpert pairs. RadFact and RadMatch use local Qwen3.8-27B-FP8 at temperatures 0 and 1, respectively (Section~\ref{sec:implementation}).}
\label{tab:main}
\setlength{\tabcolsep}{8pt}
\begin{tabular}{lcccc}
\toprule
\multicolumn{5}{c}{Panel A: complete expert datasets}\\
 & \multicolumn{2}{c}{RadEvalX ($n=100$)} & \multicolumn{2}{c}{RadEvalExpert ($n=624$)}\\
Method & Total & Significant & Total & Significant\\\midrule
Jev--OneQ & \shortstack{0.573\\\scriptsize [0.470, 0.661]} & \shortstack{0.346\\\scriptsize [0.204, 0.475]} & \shortstack{0.398\\\scriptsize [0.339, 0.455]} & \shortstack{0.352\\\scriptsize [0.287, 0.410]}\\[3pt]
Jev--Full & \shortstack{0.582\\\scriptsize [0.480, 0.675]} & \shortstack{0.343\\\scriptsize [0.202, 0.470]} & \shortstack{0.397\\\scriptsize [0.337, 0.453]} & \shortstack{0.350\\\scriptsize [0.286, 0.408]}\\[3pt]
Open NLI & \shortstack{0.337\\\scriptsize [0.205, 0.457]} & \shortstack{0.162\\\scriptsize [-0.005, 0.310]} & \shortstack{0.323\\\scriptsize [0.260, 0.381]} & \shortstack{0.276\\\scriptsize [0.207, 0.339]}\\[3pt]
Finding count & \shortstack{0.288\\\scriptsize [0.145, 0.411]} & \shortstack{0.123\\\scriptsize [-0.032, 0.275]} & \shortstack{0.261\\\scriptsize [0.189, 0.324]} & \shortstack{0.209\\\scriptsize [0.137, 0.274]}\\[3pt]
GREEN & \shortstack{0.448\\\scriptsize [0.327, 0.562]} & \shortstack{0.347\\\scriptsize [0.194, 0.487]} & \shortstack{0.180\\\scriptsize [0.101, 0.254]} & \shortstack{0.193\\\scriptsize [0.116, 0.267]}\\[3pt]
RadCliQ & \shortstack{0.375\\\scriptsize [0.244, 0.491]} & \shortstack{0.277\\\scriptsize [0.125, 0.420]} & \shortstack{0.143\\\scriptsize [0.076, 0.208]} & \shortstack{0.134\\\scriptsize [0.070, 0.199]}\\[3pt]
Temporal F1 & \shortstack{0.410\\\scriptsize [0.275, 0.531]} & \shortstack{0.234\\\scriptsize [0.060, 0.386]} & \shortstack{0.257\\\scriptsize [0.187, 0.325]} & \shortstack{0.283\\\scriptsize [0.212, 0.351]}\\[3pt]
RadGraph F1 & \shortstack{0.227\\\scriptsize [0.103, 0.344]} & \shortstack{0.129\\\scriptsize [-0.024, 0.263]} & \shortstack{0.014\\\scriptsize [-0.062, 0.081]} & \shortstack{0.026\\\scriptsize [-0.053, 0.098]}\\[3pt]
BERTScore & \shortstack{0.107\\\scriptsize [-0.042, 0.257]} & \shortstack{0.000\\\scriptsize [-0.162, 0.146]} & \shortstack{0.034\\\scriptsize [-0.043, 0.112]} & \shortstack{0.039\\\scriptsize [-0.036, 0.115]}\\[3pt]
RadFact (local 27B) & \shortstack{0.477\\\scriptsize [0.355, 0.581]} & \shortstack{0.398\\\scriptsize [0.265, 0.526]} & \multicolumn{2}{c}{See Panel B}\\[3pt]
RadMatch (local 27B) & \shortstack{0.580\\\scriptsize [0.473, 0.679]} & \shortstack{0.491\\\scriptsize [0.365, 0.613]} & \multicolumn{2}{c}{See Panel B}\\[3pt]
\bottomrule\end{tabular}
\par\medskip
\begin{tabular}{lcc}
\toprule
\multicolumn{3}{c}{Panel B: common RadEvalExpert subset ($n=200$)}\\
Method & Total & Significant\\\midrule
Jev--OneQ & \shortstack{0.322\\\scriptsize [0.219, 0.415]} & \shortstack{0.259\\\scriptsize [0.152, 0.354]}\\[3pt]
Jev--Full & \shortstack{0.320\\\scriptsize [0.216, 0.417]} & \shortstack{0.258\\\scriptsize [0.153, 0.356]}\\[3pt]
Open NLI & \shortstack{0.294\\\scriptsize [0.198, 0.393]} & \shortstack{0.230\\\scriptsize [0.126, 0.336]}\\[3pt]
RadFact (local 27B) & \shortstack{0.151\\\scriptsize [0.023, 0.281]} & \shortstack{0.186\\\scriptsize [0.058, 0.309]}\\[3pt]
RadMatch (local 27B) & \shortstack{0.453\\\scriptsize [0.348, 0.549]} & \shortstack{0.485\\\scriptsize [0.385, 0.580]}\\[3pt]
\bottomrule\end{tabular}
\end{table}

%% file: tables/table2.tex
\begin{table}[t]\centering\small
\caption{Error scope changes the ordering of Jev and open NLI on ReXErr. Each row uses the same 10,790 unchanged negative sentences. Positive counts differ by scope. Differences are Jev minus NLI; intervals use study-block bootstrap.}
\label{tab:scope}
\begin{tabular}{lrrrr}\toprule
Positive error scope & $n_+$ & Jev AUROC & NLI AUROC & Difference [95\% CI]\\\midrule
All error types & 8,723 & 0.8896 & 0.8969 & -0.0073 [-0.011, -0.003]\\
Factual error types & 7,085 & 0.9680 & 0.9614 & +0.0066 [+0.004, +0.009]\\
Language perturbations & 1,638 & 0.5505 & 0.6179 & -0.0674 [-0.083, -0.052]\\
\bottomrule\end{tabular}\end{table}

%% file: tables/repeatability.tex
\begin{table}[t]\centering\small
\caption{Repeatability on the same 50 RadEvalExpert pairs, with five runs per setting. Identical is the fraction of pairs with exactly the same score in all five runs. Rank consistency is the median Kendall $\tau_b$ across the ten pairs of runs. Correlation SD measures variation in $\tau_b$ against expert total-error counts. Jev repeats uncached judgments with L1 decomposition fixed; baseline pipelines are rerun. RadFact uses logical F1; RadMatch uses actionable-error counts.}
\label{tab:repeatability}
\begin{tabular}{llrrr}\toprule
Evaluator & Setting & Identical & Rank consistency & Correlation SD\\\midrule
Jev--OneQ & Fixed L1 & 0\% & 0.990 & 0.00381\\
Jev--Full & Fixed L1 & 0\% & 0.986 & 0.00126\\
GREEN & Greedy & 100\% & 1.000 & 0.00000\\
RadFact & Temperature 0 & 88\% & 0.928 & 0.02182\\
RadMatch & Temperature 0 & 82\% & 0.965 & 0.00975\\
RadMatch & Temperature 1 & 10\% & 0.662 & 0.03582\\
\bottomrule\end{tabular}\end{table}

%% file: sections/06_discussion.tex
\section{Discussion}
Jev combines support judgments in both directions to score unsupported
candidate statements and missing reference findings. The resulting report
scores track expert-assessed error burden. This gives researchers a low-cost way to score collections of
generated reports, with individual support judgments available for inspection.

OneQ retains expert agreement close to Full while reducing judgment API input
cost by 43--45\%. Additional questions and filtering offer no consistent gain,
and the matched NLI comparison shows that the choice of judge matters within
this simple configuration. The result is a practical starting point for
report-generator evaluation: extract statements, ask one support question,
and aggregate discrepancies in both directions.

The choice also depends on the evaluation target. On RadEvalX, Jev and local
RadMatch have similar total-error correlations, while RadMatch agrees more
closely with clinically significant errors. On the shared RadEvalExpert
subset, RadMatch leads on both targets. Researchers prioritizing significant
errors therefore have a reason to use the more elaborate evaluator. Jev's
contribution is an inexpensive judgment component that detects medically
meaningful discrepancies through a simple support question.

Discrepancy counts retain
information about report size; normalized scores describe the proportion of
content in agreement. An evaluator that tolerates a harmless spelling change
may suit factuality assessment while missing an error in a writing-quality
test. Finding-count baselines, comparisons on shared samples, and breakdowns
by error scope help connect a metric's benchmark score to its intended use.
Where multiple candidates describe the same examination, within-study ordering
adds a direct view of choosing between those candidates.

Jev's aggregate rankings repeat closely across runs, while GREEN returns
identical scores.
RadMatch's rank consistency rises from 0.662 to 0.965 when sampling is disabled
on the same inputs. Decoding settings affect repeatability, while calibration
improves probability estimates for selecting statements for further inspection.

\subsection{Limitations and future work}
Our evaluation measures agreement with reference text. Reference reports can
contain errors or omit relevant findings; establishing correctness against
the image and validating clinical use require separate studies. Report-level
category scores do not yet reliably distinguish location, severity, and
false-finding errors. Probability thresholds benefit from calibration on the
intended data, and language-quality assessment requires coverage beyond
factual support.

The study covers one Jev version, small expert benchmarks, and controlled
injected errors. RadEvalX uses one generator and RadCliQ-stratified sampling,
with 80 abnormal and 20 normal reports. Broader generator and case distributions
would test generality. Sentence-level detection and calibration use Full's
support outputs; direct evaluation of OneQ's local probabilities is a further
test. Repeating decomposition would extend Jev's fixed-decomposition
repeatability study to the full pipeline. Finally, matched end-to-end cost
and timing measurements would complement the measured judgment costs and
three-finding API demonstration when comparing deployment options.

%% file: sections/07_conclusion.tex
\section{Conclusion}
Evaluating a generated radiology report requires identifying what it changes,
adds, and leaves out relative to the reference. Jev scores these discrepancies
through simple statement-level support judgments combined in both directions.
Its single-question configuration achieves competitive agreement with expert
error counts at a judgment API cost below three cents per hundred report
pairs, excluding local decomposition. Controlled-error tests demonstrate
sensitivity to several medically meaningful changes, and repeated calls
produce closely agreeing report rankings. One support question per statement
yields expert agreement close to the seven-question design at 43--45\% lower
judgment cost. This simple Jev configuration offers a low-cost way to evaluate
factual differences in generated reports.

%% file: sections/10_availability.tex
\section*{Data and Code Availability}
RadEvalX and ReXErr-v1 (both version 1.0.0) are available through
PhysioNet~\citep{radevalx_iu,rexerr}; RadEvalExpert and RaTE-Eval are available
from their respective dataset distributions~\citep{radeval,ratescore},
subject to each dataset's terms of use.
Exact questions and implementation details appear in
Appendix~\ref{app:implementation}.
The analysis scripts and aggregate results are not yet publicly available;
we are preparing them for release.

%% file: sections/08_appendix.tex
\section{Data Processing and Supplementary Tasks}
\label{app:data}
RadEvalX uses the consensus of two radiologists over eight error categories.
We sum significant and insignificant counts for the total-error target.
Blank category cells encode zero: this convention exactly reproduces all
16 category totals for the 30 high-error reports in the original paper's
Appendix D~\citep{radeval_original}. The 100 annotated report IDs join
one-to-one to the released 590-row metric file; the remaining rows have no
expert labels and are excluded. Our main baseline scores are recomputed
with the same implementations used for RadEvalExpert.

There are 100 distinct report IDs but 96 distinct reference texts and 78
candidate texts. We retain report IDs as the sampling unit; identical text
alone does not identify a shared study. Each ID has only one candidate,
so within-study candidate ranking is unavailable on RadEvalX.
Table~\ref{tab:radevalx_sensitivity} removes the two uncertainty categories
from the labels while keeping predictions fixed.
\input{tables/radevalx_sensitivity}

RadEvalExpert preserves its total and significant-error labels.
The study identifier is the resampling unit, keeping candidates with their
shared reference together. ReXErr sentence inputs contain one candidate
sentence and its complete original report. We exclude the dataset's neutral
rewrites from the main binary-error analysis. Unchanged sentences form the
negative class in every error-scope comparison.

The factual-error group contains false negation, false prediction, changed
location, changed severity, changed measurement, added medical device,
changed device position, and changed device name. The language group contains
added typos, homophone substitutions, and repetition. This is an exploratory
grouping by the dataset's intervention labels. We have not manually adjudicated
the medical meaning of every edit. The two groups share unchanged negatives
but use disjoint positive examples.

RaTE-Eval contributes 440 sentence pairs, 370 paragraph pairs, and 847
synthetic synonym--opposite comparisons. Each synthetic comparison produces
two reference--candidate pairs, giving 1,694 pairs in total.
The distributed synthetic files contain rewrites and opposites without their
original sentences. We reconstruct references from the sentence-task pool
using text similarity and one-to-one assignment. Full gives the synonym a lower error
score in 97.3\% of these comparisons and a higher soft F1 in 99.9\%.
Excluding the seven comparisons below the preprocessing confidence
threshold of 0.20 leaves 840 comparisons. The corresponding rates are
97.5\% and 100.0\%. These results describe the reconstructed task;
the expert benchmarks and ReXErr analyses use their supplied references.

Table~\ref{tab:rate} reports the other RaTE-Eval tasks. The normalized
Jev count correlates with sentence error ratios at 0.398 and paragraph
quality ratings at 0.424. RadCliQ reaches 0.487 on paragraph ratings,
showing that broader report quality remains a distinct evaluation target.
\input{tables/rate}

\section{Implementation and Exact Questions}
\label{app:implementation}
Report segmentation uses spaCy with repairs for abbreviations, measurements,
and numbered lists. Template headings and empty units are removed.
L1 passes each remaining sentence to the atomic-decomposition prompt below.
The local decomposition model is Qwen3.5-9B, with temperature zero.
Its invalid JSON outputs fall back to the input sentence.

Jev receives the claim and opposing report as structured state fields.
Each statement--report comparison uses one request: Full includes all seven
questions, while OneQ includes only the support question. Both configurations
check statements in both directions. Repeated-call experiments obtain fresh
judgments for the same inputs.

The default weights in Eq.~\ref{eq:error} are one. The weight grid varies
candidate-side not-addressed and reference-side not-addressed terms over
$\{0.5,1,2\}^2$, leaving candidate contradiction weight fixed at one.
Significance weighting multiplies each discrepancy term by the associated
expected significance score divided by two. Category scores distribute
candidate discrepancy mass according to the error-type probabilities;
reference not-addressed mass supplies the omission category.
These diagnostics use Full's additional outputs.

\input{sections/09_prompts}

\subsection{Filtering and shortened criteria}
The filtering variant asks, ``This sentence is about the same anatomical
entity as the claim,'' for every report sentence. It retains sentences whose
probability exceeds 0.3 and passes their concatenation to Full. The shortened
support criteria are ``The report states the same finding,'' ``The report
states the opposite,'' and ``The report does not mention it.'' They replace
the three support criteria in order; the other Full questions stay the same.

\section{Additional Expert Comparisons}
\label{app:metrics}
Table~\ref{tab:all_metrics} reports 17 standard metric outputs, including
variants from the same method family. Similarity metrics are negated when
correlated with error counts. A negative directed value is retained and
indicates an association opposite to the expected direction.
All subset correlations and paired differences are recomputed on the shared valid pairs
before study resampling.
\input{tables/all_metrics}

For within-study ordering, every unordered pair of candidates contributes
$+1$, $-1$, or zero according to whether the method and expert orderings agree,
disagree, or contain a tie. The denominator includes all candidate pairs.
Full RadEvalExpert contributes 624 comparisons.
The common 200-pair RadFact/RadMatch subset spans 138 studies and contributes
69 comparisons. Each bootstrap draw sums the pair counts of the sampled
study blocks, with repeated studies contributing repeated counts.
Within-study comparisons share a reference report; candidate lengths can
still differ.

\subsection{Source, section, and modality}
Table~\ref{tab:expert_strata} separates RadEvalExpert by data source and report
section. Jev--Full has positive total-error correlation in each source,
with wider uncertainty for the smallest source. Table~\ref{tab:modality_strata}
reports the RaTE-Eval modalities with at least 20 evaluated pairs.
Performance varies with both modality and evaluation target; paragraph
quality ratings often favor BERTScore or RadCliQ. These strata show how
performance varies within the evaluated datasets.
\input{tables/expert_strata}
\input{tables/modality_strata}

\section{Complete Component Comparisons}
\label{app:ablation}
Table~\ref{tab:ablation_detail} gives the numerical comparisons shown in
Figure~\ref{fig:ablation}. Table~\ref{tab:weight_grid} covers all nine weight
combinations. The default configuration is retained throughout the main
comparisons. RadEvalX peaks at $w_h=w_o=0.5$, with $\tau_b=0.635$
and paired gain $0.053$ [0.005, 0.104]. RadEvalExpert peaks at $w_h=0.5$, $w_o=1$, with $\tau_b=0.410$.
These exploratory settings were compared on the evaluation data, so the
largest observed value is not an independently validated tuning gain.
\input{tables/ablation_detail}
\input{tables/weight_grid}

\section{Cost and Latency}
\label{app:cost}
Let $T_{v,i}$ denote the returned input-token count for configuration $v$ on
report pair $i$, summed over both directions. With $N$ pairs, workload cost is
\begin{equation}
 \overline{\mathrm{cost}}_v
 =\frac{1}{N}\sum_i T_{v,i}\frac{0.042}{10^6},\qquad
 \mathrm{saving}=1-\frac{\sum_iT_{\mathrm{OneQ},i}}
                              {\sum_iT_{\mathrm{Full},i}}.
\end{equation}
On RadEvalX, $1-5374.55/9695.17=44.56\%$.
On RadEvalExpert, $1-6151.1346/10797.1971=43.03\%$.
The resulting mean API costs are shown in Table~\ref{tab:cost}.
Full and OneQ use the same L1 decomposition, so its local cost is shared.

\begin{table}[htbp]\centering\small
\caption{API workload cost at \$0.042 per million input tokens. Cached judgments
are priced at the same rate as fresh judgments. Local GPU costs are excluded.}
\label{tab:cost}
\begin{tabular}{llrr}\toprule
Dataset & Configuration & Input tokens/pair & USD/pair\\\midrule
RadEvalX & Full & 9,695.17 & 0.00040720\\
         & OneQ & 5,374.55 & 0.00022573\\
RadEvalExpert & Full & 10,797.20 & 0.00045348\\
             & OneQ & 6,151.13 & 0.00025835\\\bottomrule
\end{tabular}\end{table}

Full on RadEvalExpert has median request latency 433.6\,ms
and 95th-percentile latency 548.4\,ms. Its median pair-level judgment time
is 0.71\,s with concurrent statement requests. Batch completion took
10.2 minutes, including scheduling and rate limits. Request-latency
measurements exclude
local cache hits, whereas the workload-cost comparison above includes them.
No matched cross-method timing of the full report-evaluation pipeline is included.

\subsection{Three-finding API demonstration}
\label{app:api-demo}
Figure~\ref{fig:api-demo} uses three constructed claims fixed before execution.
Each method receives the same complete reference and the support question and
category definitions in Appendix~\ref{app:prompts}. We send one request per
finding, with concurrency three, for five rounds in alternating method order.
There is no local response cache or warmup. We show first-round outputs and
summarize all five rounds; all 30 classifications match the expected labels.
Timing covers the judgment requests and excludes decomposition.

Jev uses version \texttt{jev-1.13.0} with OneQ. Luna responses report
\texttt{gpt-5.6-luna}, standard service, and zero reasoning tokens. Its requests
set reasoning effort to \texttt{none}, temperature to zero, JSON output,
and a 64-token output limit. Completion time includes communication and
service overhead. The five-run ranges are 0.607--1.212\,s for
Jev and 2.869--4.301\,s for Luna.

At the September 22, 2026 standard API prices~\citep{jev,openai_pricing},
Jev's 1,201 input tokens per round cost
$1201\times0.042/10^6=\$0.000050442$.
Luna uses 686 input tokens and 40 output tokens per round, with no cached
input or cache-write tokens. Its cost is
$(686\times0.20+40\times1.20)/10^6=\$0.00018520$.

\section{Error Types and Language Stress Tests}
\label{app:errors}
Table~\ref{tab:error_types} gives the complete sentence-level error profile.
The 131-item constructed stress set probes nine forms of negation,
uncertainty, and clinical attributes. Jev assigns the expected three-way
support label in 125 cases (95.4\%) in the original run.
After error review, we corrected three device-status labels and simplified
one compound claim, obtaining 129 correct cases (98.5\%).
The corrections followed the support criteria: an unmentioned device is
unaddressed, and each claim should state one finding.
Both results are retained because these revisions followed inspection of
model outputs. On the revised set, Jev answers all 15 laterality cases,
22 direct-negation cases, and 16 uncertainty cases correctly.
These are small constructed sets, rather than estimates of clinical
sensitivity for naturally occurring errors.
\input{tables/error_types}

Two cases fail. First, a double-negation statement, ``There is no evidence
to suggest the absence of a left pleural effusion,'' is labeled contradicted
against a report stating a left effusion. Second, a claim about a femur
fracture is labeled contradicted rather than unaddressed against a chest
report containing ``No acute osseous abnormality.'' The latter illustrates
how broad negative language can blur contradiction and noncoverage.

\begin{figure}[t]
\centering\includegraphics[width=\linewidth]{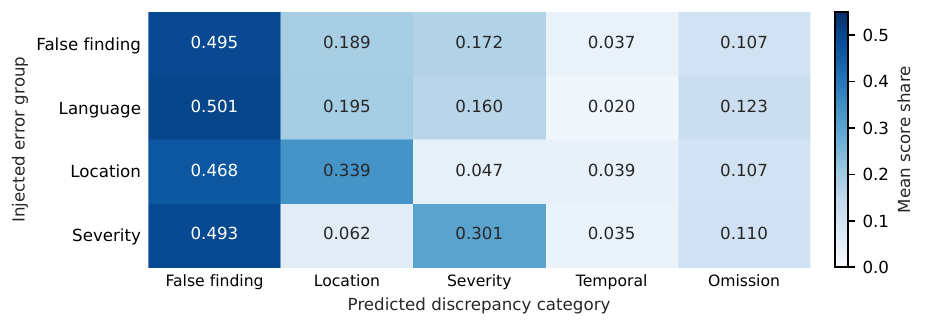}
\caption{Report-level category allocation on 2,000 sampled ReXErr pairs.
Rows group the injected errors; cells show mean proportions of predicted
category score. The omission component uses the reverse direction.
This diagnostic measures score allocation, not classification accuracy.}
\label{fig:categories}
\end{figure}

Report-level category attribution is weaker than binary error detection.
Figure~\ref{fig:categories} shows each predicted category's average share of
total category score, grouped by injected error class. For location errors,
the false-finding share is 0.468 and the location share is 0.339.
For severity errors, the corresponding shares are 0.493 and 0.301.
The broad false-finding category dominates even when a more specific category
is expected. The matrix summarizes score allocation, not a finding-matched
confusion matrix. These results do not support treating the category scores
as reliable report-level error explanations.

\section{Calibration and Repeated Calls}
\label{app:calibration}
\begin{figure}[htbp]
\centering\includegraphics[width=\linewidth]{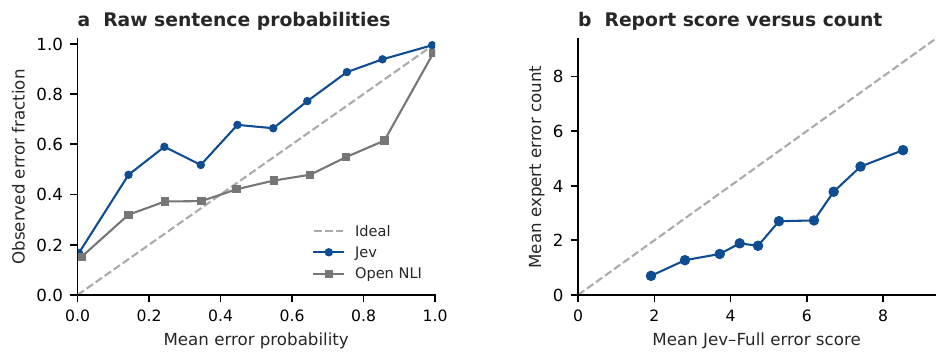}
\caption{Calibration diagnostics. (a) Raw sentence error probabilities on
all 19,513 ReXErr examples, using ten equal-width probability bins.
(b) Jev--Full score bins versus mean expert errors on RadEvalX, using
ten score quantiles. Dashed lines indicate
perfect calibration or equality with expert counts. Neither plot uses the
post-hoc calibrated holdout probabilities.}\label{fig:calibration}
\end{figure}

The raw ReXErr sentence experiment has 19,513 examples and an error prevalence
of 44.7\%. Jev's mean error probability is 0.3278; NLI's is 0.3755.
Their Brier scores are 0.1235 and 0.1192, respectively.
Figure~\ref{fig:calibration} shows the raw reliability curves.
Bin populations are uneven: Jev assigns 12,377 examples to the lowest bin
and 5,761 to the highest. The middle bins contain 113--301 examples each.
The observed error fraction in Jev's lowest bin is 0.1672 despite a mean
predicted probability of 0.0037.

Post-hoc calibration uses a fixed, equal split of studies into development
and evaluation sets. This gives 9,562
development examples and 9,951 holdout examples, without study overlap.
Logistic calibration fits a sigmoid to raw error probabilities; isotonic
regression fits a monotone map with out-of-range values clipped.
Table~\ref{tab:calibration} compares raw and calibrated outputs on the same
holdout. AUROC remains nearly unchanged in this experiment; isotonic maps
can introduce ties. These results concern the Full sentence configuration
and this error mixture.

\begin{table}[htbp]\centering\small
\caption{Probability calibration on the same 9,951-example holdout. Fits use
9,562 examples from disjoint studies. Full-set raw ECE is reported separately
in the main text.}\label{tab:calibration}
\begin{tabular}{llrrr}\toprule
Judge & Calibration & ECE & Brier & AUROC\\\midrule
Jev & Raw & 0.1235 & 0.1269 & 0.8858\\
    & Logistic & 0.0129 & 0.1061 & 0.8858\\
    & Isotonic & 0.0077 & 0.1034 & 0.8858\\
Open NLI & Raw & 0.1072 & 0.1224 & 0.8938\\
    & Logistic & 0.0185 & 0.1115 & 0.8938\\
    & Isotonic & 0.0063 & 0.1064 & 0.8937\\\bottomrule
\end{tabular}\end{table}

\subsection{Repeatability}
\label{app:repeatability}
The RadEvalX experiment randomly selects 50 report pairs and holds
615 extracted findings fixed. Five uncached runs each of Full and OneQ
produce 6,150 judgments, all using \texttt{jev-1.13.0}.
The sample standard deviations in correlation with expert total-error counts
are 0.00745 and 0.00737, respectively. Table~\ref{tab:radevalx_repeats}
summarizes ranking and score variation on this independent dataset.
\input{tables/radevalx_repeats}

The matched repeatability experiment evaluates 50 RadEvalExpert pairs from 48
studies, with five runs per setting. These are a fixed selection of 50 pairs
from the 200-pair subset. GREEN uses greedy decoding; RadFact uses its
default temperature zero. RadMatch uses the same local Qwen3.8-27B-FP8 backend
at temperature zero and at its default temperature one.

Jev--Full and Jev--OneQ use the same 616 L1 statements: 297 candidate
statements and 319 reference statements. Five rounds alternate Full and OneQ,
with 616 uncached judgments per configuration and round, all using
\texttt{jev-1.13.0}.

Table~\ref{tab:repeatability} counts a pair as identical when its score is
exactly equal across all five runs. Each of the six settings evaluates
the same 50 pairs five times.
Between-run rank consistency is the median of all ten pairwise Kendall
correlations. Correlation SD uses the sample standard deviation across five
correlations with expert total-error counts. Mean expert correlations are
0.202 for OneQ, 0.211 for Full, 0.221 for GREEN,
0.221 for RadFact, 0.468 for RadMatch at temperature zero,
and 0.400 for RadMatch at temperature one.

Jev's matched experiment holds decomposition fixed and measures API judgment
repeats; the baselines repeat their local evaluation pipelines. Every row in
Table~\ref{tab:repeatability} uses the same pairs and repetition count.
The RadEvalX experiment above provides an additional check on both configurations.

\input{tables/repeatability_fallback}
Disabling sampling improves scoring stability while validation failures
remain: temperature zero triggers fallback in 17 of 250 evaluations,
compared with nine at temperature one (Table~\ref{tab:repeatability-fallback}).
Repeated scores and output validation capture different aspects of
operational reliability.

%% file: tables/radevalx_sensitivity.tex
\begin{table}[htbp]\centering\small
\caption{RadEvalX label-scope sensitivity. The primary labels count all eight error categories; the six-category target excludes added and omitted uncertainty. Predictions stay fixed. Values are directed Kendall $\tau_b$.}
\label{tab:radevalx_sensitivity}
\begin{tabular}{lrrrr}\toprule
Method & Total (8) & Total (6) & Significant (8) & Significant (6)\\\midrule
Jev--Full & 0.582 & 0.577 & 0.343 & 0.339\\
Jev--OneQ & 0.573 & 0.568 & 0.346 & 0.343\\
Jev--Full soft F1 & 0.561 & 0.551 & 0.429 & 0.424\\
Open NLI & 0.337 & 0.335 & 0.162 & 0.162\\
Finding count & 0.288 & 0.295 & 0.123 & 0.126\\
GREEN & 0.448 & 0.442 & 0.347 & 0.346\\
RadMatch (local 27B) & 0.580 & 0.578 & 0.491 & 0.490\\
\bottomrule\end{tabular}\end{table}

%% file: tables/rate.tex
\begin{table}[htbp]\centering\small
\caption{Supplementary RaTE-Eval correlations. Sentence labels are error ratios (lower is better); paragraph labels are quality ratings (higher is better). Signs follow these targets. Brackets are study-bootstrap 95\% intervals.}
\label{tab:rate}
\begin{tabular}{lcc}\toprule
Method & Sentence ($n=440$) & Paragraph ($n=370$)\\\midrule
Jev--Full normalized count & 0.398 [0.332, 0.461] & 0.424 [0.353, 0.491]\\
Jev--Full soft F1 & 0.398 [0.331, 0.461] & 0.369 [0.288, 0.437]\\
Open NLI normalized count & 0.218 [0.154, 0.289] & 0.328 [0.251, 0.401]\\
BERTScore & 0.286 [0.218, 0.348] & 0.474 [0.411, 0.540]\\
RadCliQ & 0.324 [0.260, 0.382] & 0.487 [0.425, 0.553]\\
\bottomrule\end{tabular}\end{table}

%% file: sections/09_prompts.tex
\subsection{Support question and Full extensions}
\label{app:prompts}
OneQ sends the support entry below. Full sends all seven entries in one request. The state contains two text fields: \texttt{claim} and \texttt{report}.
\paragraph{support (choice).}
Does the report support the claim about the imaging finding?
\begin{description}
\item[supported] The report states the same finding with compatible presence, location, and severity
\item[contradicted] The report states the opposite, or the same finding with incompatible presence, laterality, or severity
\item[not\_addressed] The report does not mention this finding at all
\end{description}
\paragraph{claim\_negated (noul).}
The claim states that the finding is ABSENT (for example 'no', 'without', 'clear of', 'free of')
\paragraph{claim\_hedged (noul).}
The claim expresses uncertainty about the finding (for example 'possible', 'cannot exclude', 'may represent', 'questionable')
\paragraph{report\_negated (noul).}
The report states that the same finding described in the claim is ABSENT
\paragraph{report\_hedged (noul).}
The report expresses uncertainty about the same finding described in the claim
\paragraph{error\_type (choice).}
If the claim disagrees with the report, what kind of discrepancy is it?
\begin{description}
\item[false\_finding] The claim asserts a finding that the report says is absent or never mentions
\item[wrong\_location] Same finding, but a different anatomical location or laterality
\item[wrong\_severity] Same finding, but a different severity, size, or extent
\item[wrong\_temporal] Same finding, but a different comparison to prior imaging (new, unchanged, improved, worsened)
\item[none] The claim does not disagree with the report
\end{description}
\paragraph{significance (score).}
Clinical significance of the finding described in the claim, if the claim were wrong
\begin{enumerate}
\item Incidental or normal-variant statement with no effect on management
\item Would change follow-up imaging or outpatient management but not immediate care
\item Would change immediate management or is potentially life-threatening
\end{enumerate}
\subsection{Atomic decomposition prompt}
\begin{quote}\small
You split one sentence from a radiology report into atomic findings.\par 
Rules:\par 
1. Each output item states exactly one finding about one anatomical entity.\par 
2. Preserve negation and uncertainty words exactly as written ("no", "without", "cannot exclude", "possible", "unchanged").\par 
3. Preserve laterality, severity, size, and comparison-to-prior phrases inside the item they belong to.\par 
4. Do not add, infer, or normalize anything. Do not change tense.\par 
5. If the sentence already contains one finding, return it unchanged.\par 
6. Output a JSON list of strings and nothing else.
\end{quote}

%% file: tables/all_metrics.tex
\begin{table}[t]\centering\small
\caption{All 17 standard metric outputs on the complete expert datasets. Columns report directed Kendall correlation with total (T) or significant (S) errors. Variants within each method family are listed separately.}
\label{tab:all_metrics}
\begin{tabular}{lrrrr}\toprule
 & \multicolumn{2}{c}{RadEvalX} & \multicolumn{2}{c}{RadEvalExpert}\\ Metric output & Total & Significant & Total & Significant\\\midrule
bleu & 0.142 & 0.113 & -0.021 & -0.027\\
rouge1 & 0.232 & 0.172 & 0.006 & 0.005\\
rouge2 & 0.177 & 0.104 & -0.013 & -0.017\\
rougeL & 0.262 & 0.173 & 0.040 & 0.032\\
bertscore & 0.107 & 0.000 & 0.034 & 0.039\\
f1chexbert\_sample\_acc\_5 & 0.177 & 0.277 & 0.172 & 0.143\\
f1chexbert\_sample\_acc\_all & -0.031 & -0.159 & -0.167 & -0.168\\
radgraph\_simple & 0.234 & 0.086 & 0.001 & 0.009\\
radgraph\_partial & 0.225 & 0.089 & 0.002 & 0.010\\
radgraph\_complete & 0.227 & 0.129 & 0.014 & 0.026\\
ratescore & 0.290 & 0.177 & 0.073 & 0.068\\
srrbert\_weighted\_f1 & 0.332 & 0.261 & 0.139 & 0.150\\
srrbert\_weighted\_precision & 0.332 & 0.261 & 0.137 & 0.148\\
srrbert\_weighted\_recall & 0.332 & 0.261 & 0.137 & 0.148\\
temporal\_f1 & 0.410 & 0.234 & 0.257 & 0.283\\
radcliq\_v1 & 0.375 & 0.277 & 0.143 & 0.134\\
green & 0.448 & 0.347 & 0.180 & 0.193\\
\bottomrule\end{tabular}\end{table}

%% file: tables/expert_strata.tex
\begin{table}[htbp]\centering\small
\caption{RadEvalExpert total-error correlation by source and report section. Each row uses the same pairs for the three methods. Source and section partitions overlap. Intervals resample studies within each stratum.}
\label{tab:expert_strata}
\begin{tabular}{lrrrr}\toprule
Stratum & $n$ & Jev--Full & Finding count & GREEN\\\midrule
CheXpert Plus & 96 & 0.323 [0.143, 0.483] & 0.191 [-0.017, 0.351] & 0.171 [-0.017, 0.365]\\
MIMIC-CXR & 306 & 0.410 [0.335, 0.495] & 0.221 [0.137, 0.338] & 0.182 [0.079, 0.289]\\
ReXGradient & 222 & 0.396 [0.300, 0.482] & 0.341 [0.250, 0.440] & 0.257 [0.137, 0.369]\\
Findings section & 444 & 0.426 [0.353, 0.483] & 0.242 [0.165, 0.318] & 0.259 [0.159, 0.342]\\
Impression section & 180 & 0.261 [0.126, 0.361] & 0.185 [0.053, 0.314] & 0.068 [-0.086, 0.179]\\
\bottomrule\end{tabular}\end{table}

%% file: tables/modality_strata.tex
\begin{table}[htbp]\centering\small
\caption{RaTE-Eval correlation by modality, for strata with at least 20 pairs. Jev uses Full normalized error count. Similarity/error signs follow each task\textquotesingle{}s label direction. Brackets give 95\% study-bootstrap intervals.}
\label{tab:modality_strata}
\begin{tabular}{lrrrr}\toprule
Task / modality & $n$ & Jev--Full & BERTScore & RadCliQ\\\midrule
Sentence / CT & 89 & 0.386 [0.237, 0.526] & 0.235 [0.078, 0.381] & 0.312 [0.162, 0.459]\\
Sentence / CTA & 51 & 0.396 [0.219, 0.576] & 0.381 [0.209, 0.543] & 0.389 [0.227, 0.561]\\
Sentence / MRA & 32 & 0.385 [0.111, 0.632] & 0.241 [-0.057, 0.506] & 0.348 [0.109, 0.560]\\
Sentence / MRI & 88 & 0.390 [0.241, 0.525] & 0.325 [0.183, 0.455] & 0.277 [0.118, 0.412]\\
Sentence / MRV & 26 & 0.502 [0.131, 0.758] & 0.277 [-0.155, 0.545] & 0.284 [-0.120, 0.556]\\
Sentence / US & 40 & 0.397 [0.172, 0.619] & 0.197 [-0.025, 0.416] & 0.314 [0.128, 0.526]\\
Sentence / XR & 91 & 0.341 [0.198, 0.473] & 0.245 [0.083, 0.384] & 0.289 [0.140, 0.418]\\
Paragraph / CT & 62 & 0.465 [0.284, 0.625] & 0.493 [0.351, 0.637] & 0.462 [0.304, 0.624]\\
Paragraph / CTA & 64 & 0.359 [0.187, 0.526] & 0.457 [0.273, 0.613] & 0.479 [0.315, 0.612]\\
Paragraph / MRA & 38 & 0.363 [0.137, 0.539] & 0.390 [0.158, 0.608] & 0.433 [0.155, 0.645]\\
Paragraph / MRI & 84 & 0.383 [0.238, 0.509] & 0.422 [0.267, 0.563] & 0.408 [0.243, 0.559]\\
Paragraph / US & 76 & 0.429 [0.259, 0.580] & 0.573 [0.458, 0.683] & 0.557 [0.425, 0.680]\\
Paragraph / XR & 36 & 0.548 [0.325, 0.735] & 0.521 [0.273, 0.721] & 0.595 [0.379, 0.755]\\
\bottomrule\end{tabular}\end{table}

%% file: tables/ablation_detail.tex
\begin{table}[htbp]\centering\small
\caption{Paired changes in total-error $\tau_b$ relative to Full. These are independent variants, not successive simplifications. Cost ratios appear in Figure~\ref{fig:ablation}.}
\label{tab:ablation_detail}
\begin{tabular}{lrr}\toprule
Variant & RadEvalX & RadEvalExpert\\\midrule
Sentence split & -0.014 [-0.086, +0.057] & -0.015 [-0.051, +0.018]\\
OneQ & -0.009 [-0.026, +0.008] & +0.001 [-0.002, +0.004]\\
Filter state & -0.023 [-0.057, +0.011] & -0.009 [-0.020, +0.001]\\
Short criteria & -0.003 [-0.020, +0.015] & +0.002 [-0.003, +0.008]\\
No negation fix & -0.002 [-0.015, +0.012] & +0.002 [-0.001, +0.005]\\
Significance weighting & -0.129 [-0.208, -0.050] & -0.021 [-0.051, +0.005]\\
\bottomrule\end{tabular}\end{table}

%% file: tables/weight_grid.tex
\begin{table}[htbp]\centering\small
\caption{Exploratory weight grid for Full. Weights multiply candidate noncoverage ($w_h$) and reference noncoverage ($w_o$); contradiction weight stays one. Columns give total-error $\tau_b$ and paired change from $(1,1)$. Intervals are unadjusted 95\% study-bootstrap intervals.}
\label{tab:weight_grid}
\begin{tabular}{rrrrrr}\toprule
$w_h$ & $w_o$ & RadEvalX $\tau_b$ & Change [95\% CI] & RadEvalExpert $\tau_b$ & Change [95\% CI]\\\midrule
0.5 & 0.5 & 0.635 & +0.053 [+0.005, +0.104] & 0.407 & +0.010 [-0.006, +0.025]\\
0.5 & 1 & 0.597 & +0.015 [-0.026, +0.056] & 0.410 & +0.013 [-0.006, +0.035]\\
0.5 & 2 & 0.496 & -0.086 [-0.159, -0.020] & 0.381 & -0.016 [-0.049, +0.017]\\
1 & 0.5 & 0.560 & -0.022 [-0.077, +0.032] & 0.371 & -0.026 [-0.047, -0.007]\\
1 & 1 & 0.582 & +0.000 [+0.000, +0.000] & 0.397 & +0.000 [+0.000, +0.000]\\
1 & 2 & 0.519 & -0.062 [-0.112, -0.018] & 0.393 & -0.004 [-0.025, +0.019]\\
2 & 0.5 & 0.450 & -0.132 [-0.205, -0.059] & 0.318 & -0.078 [-0.116, -0.044]\\
2 & 1 & 0.492 & -0.090 [-0.136, -0.045] & 0.351 & -0.046 [-0.069, -0.025]\\
2 & 2 & 0.514 & -0.068 [-0.101, -0.039] & 0.381 & -0.016 [-0.026, -0.006]\\
\bottomrule\end{tabular}\end{table}

%% file: tables/error_types.tex
\begin{table}[t]\centering\small
\caption{Jev detection by injected error type. Each row includes the same 10,790 unchanged negative sentences. This sentence experiment uses the Full question set and raw support probabilities.}
\label{tab:error_types}
\begin{tabular}{lrrrr}\toprule
Injected error & $n_+$ & AUROC & 95\% CI & Recall at 0.5\\\midrule
Add medical device & 910 & 0.9888 & [0.984, 0.993] & 0.9451\\
Change position of device & 438 & 0.9869 & [0.979, 0.992] & 0.9041\\
Change location & 950 & 0.9841 & [0.978, 0.989] & 0.8947\\
False negation & 1398 & 0.9773 & [0.972, 0.982] & 0.9185\\
Change severity & 849 & 0.9753 & [0.968, 0.981] & 0.8492\\
Change measurement & 67 & 0.9707 & [0.937, 0.996] & 0.8955\\
False prediction & 2059 & 0.9471 & [0.939, 0.955] & 0.8436\\
Change name of device & 414 & 0.9226 & [0.905, 0.941] & 0.6691\\
Change to homophone & 127 & 0.6477 & [0.601, 0.695] & 0.0945\\
Add typo & 1131 & 0.5617 & [0.549, 0.576] & 0.0292\\
Add repetition & 380 & 0.4848 & [0.469, 0.500] & 0.0026\\
\bottomrule\end{tabular}\end{table}

%% file: tables/radevalx_repeats.tex
\begin{table}[htbp]\centering\small
\caption{Repeated Jev judgments on 50 RadEvalX pairs, five uncached runs per configuration, holding 615 extracted findings fixed. Rank agreement is the median of ten between-run Kendall correlations; correlation SD is across the five correlations with total-error labels.}
\label{tab:radevalx_repeats}
\begin{tabular}{lrrr}\toprule
Configuration & Rank agreement & Correlation SD & Median score SD\\\midrule
Jev--Full & 0.965 & 0.00745 & 0.054\\
Jev--OneQ & 0.974 & 0.00737 & 0.042\\
\bottomrule\end{tabular}\end{table}

%% file: tables/repeatability_fallback.tex
\begin{table}[htbp]\centering\small
\caption{RadMatch matching-stage validation across all five runs per setting. Each setting includes 250 report-pair evaluations on the same 50 pairs. Retried counts evaluations requiring at least one retry.}
\label{tab:repeatability-fallback}
\begin{tabular}{lrr}\toprule
Setting & Validation fallback & Retried\\\midrule
Temperature 0 & 17/250 (6.8\%) & 61/250 (24.4\%)\\
Temperature 1 & 9/250 (3.6\%) & 70/250 (28.0\%)\\
\bottomrule\end{tabular}\end{table}